\documentclass[11pt]{article}
\usepackage[final]{acl} 
\usepackage{times}
\usepackage{latexsym}
\usepackage[T1]{fontenc}
\usepackage[utf8]{inputenc}
\usepackage{microtype}
\usepackage{inconsolata}
\usepackage{graphicx}
\usepackage{algorithm}
\usepackage{algpseudocode}
\usepackage{amsmath}
\usepackage{amsfonts}
\usepackage{float}
\usepackage[most]{tcolorbox}
\tcbset{
  colback=gray!5,
  colframe=black!40,
  boxrule=0.5pt,
  arc=2pt,
  left=6pt,right=6pt,top=6pt,bottom=6pt
}
\title{Interactive AI NPCs Powered by LLMs: Technical Report for the CPDC Challenge 2025}

\author{
  Yitian Huang\thanks{Work done during internship at Microsoft Research Asia.} \\
  CSSE, Shenzhen University \\
  \texttt{yitianhuang126@gmail.com} 
  \And
  Yuxuan Lei\footnotemark[1] \\
  University of Science and Technology of China \\
  \texttt{leiyuxuan@mail.ustc.edu.cn} 
  \AND
  Jianxun Lian \\
  Microsoft Research Asia \\
  \texttt{jianxun.lian@outlook.com} 
  \And
  Hao Liao \\
  CSSE, Shenzhen University \\
  \texttt{haoliao@szu.edu.cn} 
}

\begin{document}
\maketitle

\begin{abstract}
This report presents the solution and results of our team MSRA\_SC in the Commonsense Persona-Grounded Dialogue Challenge (CPDC 2025). We propose a simple yet effective framework that unifies improvements across both GPU Track and API Track. Our method centers on two key components. First, Context Engineering applies dynamic tool pruning and persona clipping for input compression, combined with post-processing techniques such as parameter normalization and function merging. Together with manually refined prompts, this design improves tool call stability, execution reliability, and role-playing guidance. Second, in the GPU Track, we further adopt GRPO training, replacing supervised fine-tuning with reinforcement learning directly optimized by reward signals. This mitigates small-sample overfitting and significantly enhances task-oriented dialogue performance. In the final evaluation, our team ranks 1st in Task 2 API, 2nd in Task 1 API, and 3rd in both Task 3 API and GPU track, demonstrating the effectiveness of our approach. Our code is publicly available at \url{https://gitlab.aicrowd.com/nikoo_yu/cpdc-2025-winning-solution}.

\end{abstract}

\section{Introduction}
Non-Player Character (NPC) dialogue is crucial for immersion and interactivity in modern games \cite{wang2025coser,samuel2024personagym,tu2023characterchat,shao2023character}. If NPCs can generate fluent responses grounded in persona, world knowledge, and player actions, they not only improve user experience but also enhance the interpretability of task execution. However, current NPC dialogue systems often suffer from rigidity, incoherence, and weak adaptation to dynamic goals.  

To address these issues, the \textit{Commonsense Persona-Grounded Dialogue Challenge (CPDC 2025)} \footnote{https://www.aicrowd.com/challenges/commonsense-persona-grounded-dialogue-challenge-2025}  defines three key tasks:  
\begin{itemize}
  \item \textbf{Task 1: Task-Oriented Dialogue} – whether agents can correctly call tools to accomplish tasks.  
  \item \textbf{Task 2: Context-Aware Dialogue} – whether agents can leverage persona, worldview, and dialogue history.  
  \item \textbf{Task 3: Integration} – testing agents in more complex scenarios requiring both task execution and dialogue.  
\end{itemize}

The competition includes a GPU Track (open-source models with training) and an API Track (black-box APIs without training), posing multi-faceted challenges to participants.  

In this work, our team MSRA\_SC proposes a simple but effective method based on two main improvements:  
\begin{enumerate}
  \item \textbf{Context Engineering} – dynamic tool pruning and persona clipping for input compression, combined with post-processing for parameter normalization and function merging. This improves dialogue stability and execution reliability across both tracks. We also manually refine and validates prompts, achieving optimal templates for tool call and role-playing guidance.  
  \item \textbf{GRPO Training} – in the GPU Track, we adopt pure GRPO training instead of SFT, directly optimizing results via reward signals. This approach alleviates small-sample overfitting and significantly enhances task-oriented dialogue performance.
\end{enumerate}

Our solution ranks 1st in Task 2 API, 2nd in Task 1 API, and 3rd in both Task 3 GPU and Task 3 API track, demonstrating the effectiveness and generality of our method for NPC dialogue systems, providing empirical support for building more natural and intelligent NPC dialogue systems. In this technical report, we will present a detailed description of our method and the corresponding experimental results.

\section{API Track}
\subsection{Context Engineering}
Task 1 focuses on Task-Oriented Dialogue, where models are required to accurately invoke the appropriate tools while generating responses that align with the NPC’s persona, worldview, and dialogue history. Task 2 targets Context-Aware Dialogue, which emphasizes role-playing without evaluating tool invocation capabilities. Task 3 serves as a hybrid of Task 1 and Task 2, combining both requirements.

Although defined separately, the three tasks are inherently interconnected. To address them systematically, our context engineering pipeline is organized into three stages: Preprocessing, Post-processing, and Prompt Optimization.

\subsubsection{Preprocessing}
A key constraint in the API track is the strict token budget: each turn is limited to 2,000 tokens for input and 200 tokens for output. In our experiments, we observe that directly including the full metadata (e.g., persona information and tool descriptions) often exceeds these limits. To address this, we design a dynamic pruning mechanism that adaptively reduces both the toolset and persona content.

\textbf{Adaptive Toolset Pruning.} The procedure consists of three stages:
(i) reordering tools by estimated relevance,
(ii) iteratively pruning the least relevant tools, and
(iii) truncating tool descriptions at a finer granularity.
This design ensures graceful degradation of tool information under budget constraints. Algorithm~\ref{alg:toolset-pruning} illustrates the cascaded procedure for the whole process.

\begin{algorithm}[h]
\caption{Adaptive Toolset Pruning}
\label{alg:toolset-pruning}
\begin{algorithmic}[1]
\Require Message history $M$, toolset $T$, token limit $L_{\max}$
\Ensure Optimized toolset $T_{opt}$
\State $T_{opt} \gets$ DeepCopy($T$)
\If{CalculateTokens($M, T_{opt}$) $\leq L_{\max}$}
  \State  \Return $T_{opt}$
\EndIf
\State \textbf{Stage 1: Relevance-based Reordering}
\State $q \gets$ ExtractLastUserQuery($M$)
\State $T_{opt} \gets$ SortByRelevance($T_{opt}, q$)
\If{CalculateTokens($M, T_{opt}$) $\leq L_{\max}$}
   \State \Return $T_{opt}$
\EndIf
\State \textbf{Stage 2: Iterative Pruning}
\For{$i = 1$ to $3$}
    \State RemoveLowestRankedTool($T_{opt}$)
    \If{CalculateTokens($M, T_{opt}$) $\leq L_{\max}$}
       \State \Return $T_{opt}$
    \EndIf
\EndFor
\State \textbf{Stage 3: Description Truncation}
\While{CalculateTokens($M, T_{opt}$) $> L_{\max}$}
    \For{each tool in $T_{opt}$}
        \State tool.description$\gets$Truncate($10\%$)
    \EndFor
\EndWhile
\State \Return $T_{opt}$
\end{algorithmic}
\end{algorithm}

\textbf{Hierarchical Persona Distillation.} Algorithm~\ref{alg:persona-distillation} formalizes the strategy for reducing persona-related context in dialogue scenarios. Information is pruned according to a manually defined salience hierarchy, ensuring that peripheral details (e.g., hobbies) are removed prior to central role-defining attributes (e.g., worldview or role identity).

\begin{algorithm}[h]
\caption{Hierarchical Persona Distillation}
\label{alg:persona-distillation}
\begin{algorithmic}[1]
\Require Persona components $C$, reduction level $L$
\Ensure Distilled prompt $P$
\State $C_{distilled} \gets$ DeepCopy($C$)

\State $salience\_order \gets$ [state, role, worldview, knowledge, npc\_info]
\For{$level = 1$ to $L$}
    \State $C_{distilled}[salience\_order[level]] \gets$ Truncate($C$, level)
\EndFor
\State $P \gets$ FormatPrompt($C_{distilled}$)
\State \Return $P$
\end{algorithmic}
\end{algorithm}

Overall, these preprocessing strategies reduce prompt length while preserving core context, ensuring task relevance and efficiency.

\subsubsection{Post-processing}
\label{sec:postprocessing}
The raw tool calls generated by the LLM often contain inconsistencies or redundancies. 
To address this, we implement a lightweight post-processing pipeline consisting of 
\textbf{parameter normalization} and \textbf{function merging}. 
The former ensures type correctness, canonicalizes values (e.g., mapping ``>'' to ``more than''), 
and handles operator splitting or inference from user queries. 
The latter consolidates multiple fine-grained checks into a single call, 
removes redundant or conflicting actions (e.g., avoiding selling an equipped item), 
and validates arguments against the knowledge base. 
This refinement substantially reduces invalid calls and improves execution reliability across tasks.

\subsubsection{Prompt Optimization}
Since prompt templates themselves have a substantial impact on model performance, we iteratively refine and manually validate them to derive effective formats that balance accurate tool invocation with coherent role-playing. Full prompts for Task1 and Task2 can be found in Appendix~\ref{sec:prompt_templates}.

\subsection{Experimental Results}
For the API track, the model choice is fixed by the organizers: only gpt-4o-mini \cite{hurst2024gpt} is allowed and available on the evaluation servers. As shown in Table \ref{tab:api_results}, gpt-4o-mini achieves a substantial performance improvement after applying Context Engineering.
\begin{table}[h]
\centering
\fontsize{12.3pt}{14.3pt}\selectfont
\begin{tabular}{lcc}
\hline
Model & Task1 & Task2 \\
\hline
gpt-4o-mini & 0.46 & 0.58 \\
$\text{gpt-4o-mini}_{\text{CE}}$ & 0.55 & 0.62 \\
\hline
\end{tabular}
\caption{Results on API Track (online evaluation) CE means Context Engineering.}
\label{tab:api_results}
\end{table}

\begin{figure*}[t]
    \centering
    \includegraphics[width=0.95\textwidth]{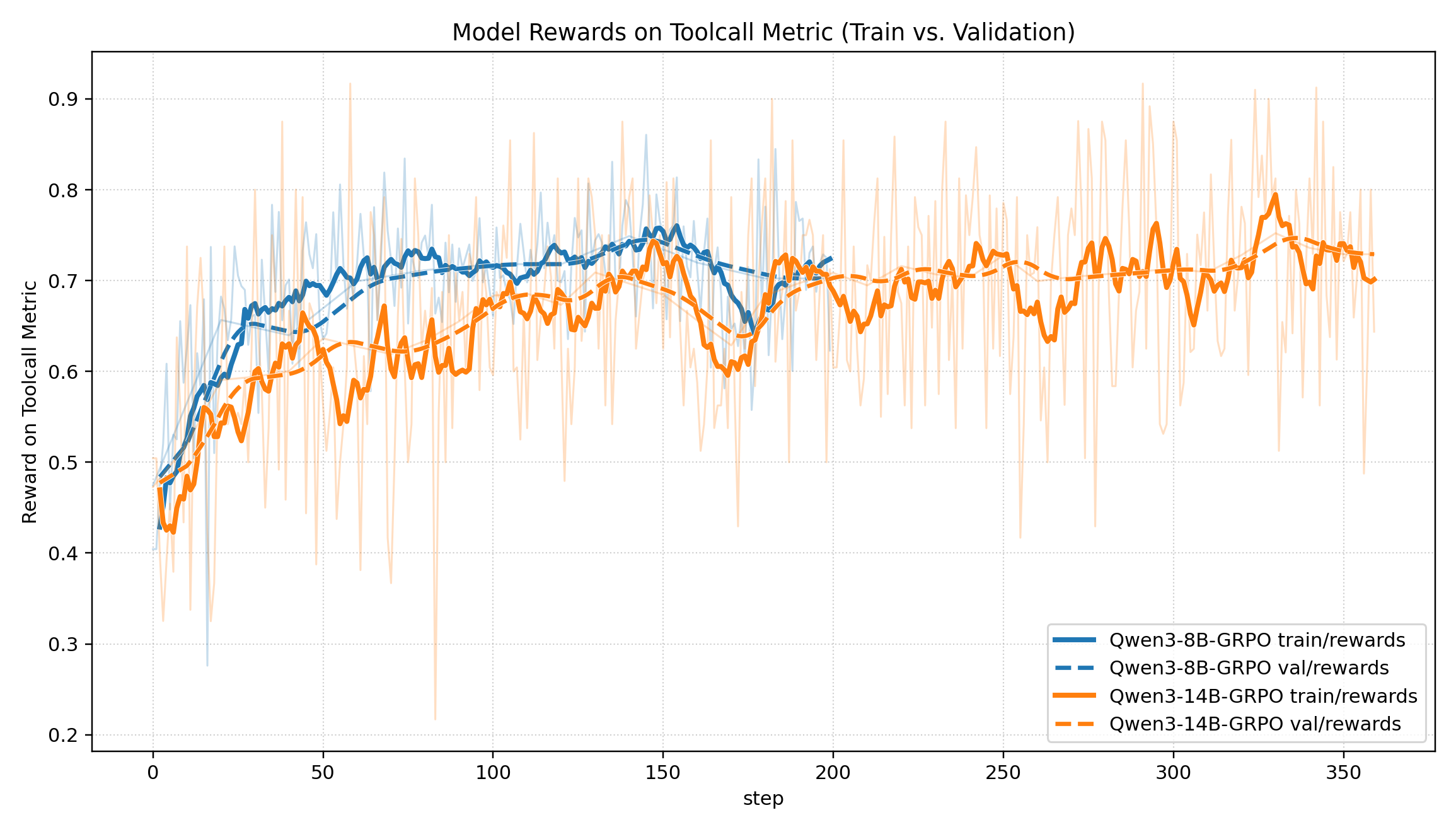}
    \caption{Tool call Train and eval curves (Qwen3-8B/14B, GRPO)}
    \label{fig:toolcall-training}
\end{figure*}
\section{GPU Track}
In the GPU track, beyond leveraging the Context Engineering method introduced in the API track, we additionally perform reinforcement learning on the limited set of training data provided by the organizers.

\subsection{GRPO Training}
We adopt \emph{Group Relative Policy Optimization} (GRPO), a PPO-style policy optimization that removes the critic/value network and computes group-relative advantages from multiple samples per prompt. For each input $x$, we draw $K$ rollouts $\{y_i\}_{i=1}^K$ from the old policy and obtain scalar rewards $\{r_i\}_{i=1}^K$. Let $\bar r=\frac{1}{K}\sum_{i=1}^K r_i$ and $\sigma_r$ be the standard deviation; the advantage is
\begin{equation}
A_i = \frac{r_i - \bar r}{\sigma_r + \epsilon}.
\end{equation}
With importance ratio
\begin{equation}
\rho_i = \frac{\pi_{\theta}(y_i \mid x)}{\pi_{\theta_{\text{old}}}(y_i \mid x)},
\end{equation}
the GRPO objective is defined as
\begin{align}
\mathcal{L}_{\text{GRPO}}
   &= - \mathbb{E}\!\left[
        \min\!\Big(
          \rho_i A_i,\;
          \operatorname{clip}(\rho_i, 1-\varepsilon, 1+\varepsilon)\, A_i
        \Big)
      \right] \notag \\
   &\quad + \beta \, \mathrm{KL}\!\left(\pi_{\theta} \,\|\, \pi_{\text{ref}}\right)
      - \alpha \, \mathcal{H}[\pi_{\theta}],
\end{align}
where $\varepsilon$ is the clipping threshold, $\beta$ controls KL regularization, and $\alpha$ is the entropy weight.
GRPO was introduced in DeepSeekMath \cite{shao2024deepseekmath} and later adopted for reasoning-oriented post-training in DeepSeek-R1 \cite{guo2025deepseekr1}.

\subsection{Reward Setup}
We use two task-specific rewards that directly mirror the official metrics.

\textbf{cpdc/tool call (Task~1).}
The reward is the tool call\_F1 between predicted and gold tool calls parsed from \texttt{<tool\_call>} blocks. 
A predicted call is correct if the function name and its arguments match one gold call (one-to-one matching).
Let $N_{\text{pred}}$, $N_{\text{gold}}$, and $N_{\text{correct}}$ be the counts of predicted, gold, and correctly matched calls, respectively. 
We compute
\begin{align}
\mathrm{Precision} &= \frac{N_{\text{correct}}}{\max(1,\,N_{\text{pred}})},\\
\mathrm{Recall} &= \frac{N_{\text{correct}}}{\max(1,\,N_{\text{gold}})},\\
r_{\text{tool}} &= \frac{2\,\mathrm{Precision}\cdot \mathrm{Recall}}{\max\big(1,\,\mathrm{Precision}+\mathrm{Recall}\big)}.
\end{align}
Edge case: if $N_{\text{pred}}{=}N_{\text{gold}}{=}0$, we set $r_{\text{tool}}{=}1.0$.

\textbf{cpdc/roleplay (Task~2).}
The reward is an LLM-as-Judge score $s\in\{0,1,2,3,4,5\}$ produced by a rubric-guided prompt, evaluating: 
(i) scenario adherence \& quest progression, 
(ii) NPC believability \& engagement, 
(iii) persona consistency, 
(iv) dialogue flow \& coherence. The full rubric-guided prompt can be found in Appendix~\ref{sec:prompt_templates}. We normalize it to $[0,1]$ by
\begin{equation}
r_{\text{dlg}}=\frac{s}{5}.
\end{equation}

\textbf{Final scalar reward.}
For single-task training we directly use $r_{\text{tool}}$ (Task~1) or $r_{\text{dlg}}$ (Task~2). 
When combining tasks (e.g., Task~3), we use a weighted sum
\begin{equation}
r_i = \eta_{\text{tool}}\, r_{\text{tool}} + \eta_{\text{dlg}}\, r_{\text{dlg}} \quad \text{with}\quad \eta_{\text{tool}},\eta_{\text{dlg}}\in[0,1].
\end{equation}
All rewards are clipped to $[0,1]$ and then group-standardized to compute $A_i$ as in Eq.~(1).
We set $(\eta_{\text{tool}}, \eta_{\text{dlg}}) = (0.5, 0.5)$ for Task~3.  The full rubric-guided prompt can be found in Appendix~\ref{sec:prompt_templates}.

\begin{table}[t]
\centering
\fontsize{12.3pt}{14.3pt}\selectfont
\begin{tabular}{lcc}
\hline
Model & Task1 & Task2 \\
\hline
Qwen3-8B & 0.36 & 0.55 \\
$\text{Qwen3-8B}_{\text{CE}}$ & 0.49 & 0.61 \\
$\text{Qwen3-8B}_{\text{CE + GRPO}}$ & 0.57 & 0.60 \\
\hline
Qwen3-14B & 0.38 & 0.56 \\
$\text{Qwen3-14B}_{\text{CE}}$ & 0.53 & 0.60 \\
$\text{Qwen3-14B}_{\text{CE + GRPO}}$ & 0.59 & 0.59 \\
\hline
\end{tabular}
\caption{Results on GPU Track (online evaluation).}
\label{tab:gpu_results}
\end{table}

\subsection{Training Details}
We train our models using the GRPO algorithm with $\gamma=1$ and $\lambda=1$. PPO clipping $\varepsilon=0.2$, entropy regularization coefficient $\alpha=0.01$, and we apply an adaptive KL penalty (target $0.1$, initial $\beta=10^{-3}$, coefficient $0.001$). For data sampling, each prompt is expanded into $K=5$ rollouts with top-$p=0.9$ and temperature $1.0$, with a maximum combined prompt/response length of 6,000 tokens. Optimization is performed with a batch size of 16, PPO mini-batch size of 16, and learning rates of $1\times 10^{-6}$ for the actor and $1\times 10^{-5}$ for the critic. We also apply weight decay ($0.01$) and gradient clipping ($1.0$). Training is distributed across 4 $\times$ A100 GPUs using FSDP \cite{zhao2023pytorch} with tensor model parallelism (size 2) and mixed precision (bfloat16). For evaluation, we follow the official CPDC online leaderboard. We primarily adopt the Qwen3 family (Qwen3-8B and Qwen3-14B) \cite{yang2025qwen3}, while preliminary experiments with other model families showed inferior results.

\subsection{Experimental Results}
As shown in Table \ref{tab:gpu_results}, applying Context Engineering alone already yields substantial improvements. Incorporating GRPO training leads to further gains, producing a qualitative boost in model performance. Figure \ref{fig:toolcall-training} illustrates the reward trajectory for tool calls, which increases steadily during the first 150 training steps. However, we observe that the improvements brought by GRPO training are primarily concentrated in Task 1, specifically in the tool-calling component: it significantly enhances the model’s ability to invoke tools, whereas the role-playing ability shows little progress. We think this discrepancy arises from biases in the LLM-as-a-judge reward signals, which may induce reward hacking—optimizing toward proxy rewards that fail to faithfully capture the quality of role-playing. We further discuss this in Section \ref{llm-as-judge}.




\section{Explorations and Lessons Learned}
\begin{table}[t]
\setlength{\belowcaptionskip}{-13.9pt}
\resizebox{1.03\linewidth}{!}{
\begin{tabular}{llccc}
\hline
Model A & Model B & Win & Loss & Draw \\
\hline
Qwen3-8B & $\text{Qwen3-8B}_{\text{GRPO}}$ & 12.5\% & 85.4\% & 2.1\% \\
Qwen2.5-7B & $\text{Qwen3-8B}_{\text{GRPO}}$ & 3.5\% & 96.0\% & 0.5\% \\
\hline
\end{tabular}
}
\caption{Comparison of LLM-as-a-judge scores on Task~2: Model A vs. Model B.}
\label{tab:task2-judge}
\end{table}
\subsection{SFT Failures}
Before investigating RL training, we first explore supervised fine-tuning (SFT). Since the official dataset contains fewer than 50 multi-turn dialogues, it is insufficient for SFT. To augment the data, we employ GPT-4o to synthesize additional samples. Specifically, given a dialogue intention, some background information, and two randomly selected demo conversations, GPT-4o is prompted to generate NPC and player information along with full multi-turn dialogues (excluding tool calls, focusing purely on role-playing). Because the synthesized data exhibits high redundancy, we apply an embedding-based deduplication process \cite{reimers2019sentence} to ensure that the pairwise similarity between samples remains below a predefined threshold. Using the combined dataset (around 30,000 dialogues in total, including both synthetic and official samples), we fine-tune Qwen2.5-14B-Instruct and Qwen3-14B. We observe a slight improvement on Qwen2.5-14B-Instruct ($+0.07$), but a performance drop on Qwen3-14B ($-0.03$). We hypothesize that this is because the synthetic data remains too homogeneous; since SFT essentially encourages the model to memorize training data, low-diversity and low-quality samples tend to cause overfitting and hinder generalization to the unseen test set.

\subsection{LLM-as-a-judge And Reward Hacking} \label{llm-as-judge}

\begin{figure}[t]
    \includegraphics[width=1.02\linewidth]{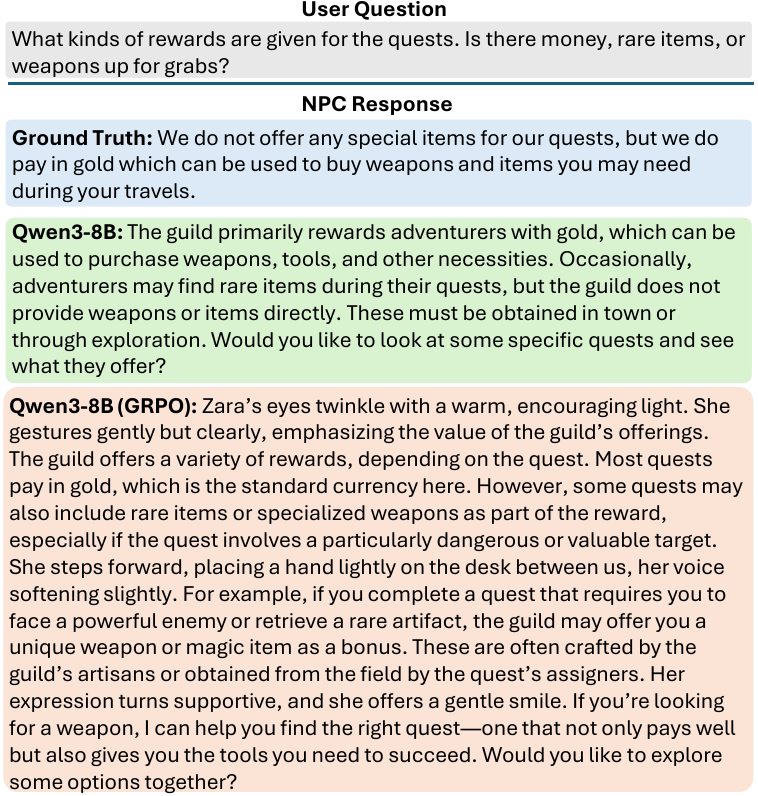}
    \caption{A case study for GRPO training with LLM-as-a-judge reward signals.}
    \label{fig:case study}
\end{figure}
During CPDC Challenge, we seek to establish a stable offline evaluation method to enable faster iteration under limited submission opportunities. For tool call task, evaluation is straightforward based on exact matching, but assessing roleplaying dialogues remains challenging. Traditional metrics such as BLEU \cite{papineni2002bleu}, Word F1 \cite{rajpurkar2016squad}, and BERTScore \cite{zhang2019bertscore} either focus on surface-level overlap or provide limited semantic alignment, leading to a substantial gap with human judgment. With the rise of large language models, LLM-as-a-judge has emerged as a widely used alternative to human evaluation \cite{gu2024survey, lei2024recexplainer, li2024llms}. We design several evaluation prompts, feeding both model responses and reference answers into GPT-4o to obtain 0–5 scores along dimensions such as scenario adherence and NPC believability. While this approach improves alignment compared to traditional metrics, offline scores still does not consistently track online performance in API track. Moreover, during GRPO training, we observe steadily increasing reward values; however, this trend does not translate into proportional improvements on the leaderboard. As shown in Table \ref{tab:task2-judge}, the GRPO-trained model demonstrates a clear advantage over its base counterpart, achieving an 85.4\% win rate under offline LLM-as-a-judge scores. Closer analysis of case studies reveals that the GRPO-trained model tends to exploit superficial prompt biases rather than genuinely improving interaction quality. For instance, as shown in Figure~\ref{fig:case study}, when asked about quest rewards, the ground-truth response provides a concise, factual answer (“gold is provided, no special items”). By contrast, our GRPO-trained model produces lengthy, dramatized role-playing outputs filled with immersive narrative details, exaggerated mentions of rare weapons or magic items, and stylistic embellishments (e.g., “Zara’s eyes twinkle with a warm, encouraging light”). While such responses increase reward scores—largely because they align with token-level preferences for in-game terminology, and vivid descriptions—they deviate from the intended factual content. This illustrates a clear case of reward hacking \cite{shihab2025detecting, hong2025cooper}, where models optimize for proxy signals in the reward function instead of faithfully improving role-playing quality. This experience highlights that while LLM-as-a-judge is more effective than classical metrics, its design must be made more robust, potentially through richer reference signals or hybrid evaluation strategies, to mitigate reward hacking and better reflect true task quality.



\section{Conclusion}
In this paper, we propose a simple yet effective approach that combines context engineering and GRPO training to improve both tool call reliability and empathetic dialogue quality. Our team rank 1st in Task 2 API, 2nd in Task 1 API, and 3rd in Task 3 API and GPU, demonstrating the effectiveness of the method. We also summarize key lessons from the competition, including the limitations of SFT and biases in automatic roleplaying evaluation. These findings lay a solid foundation for building more natural and intelligent NPC dialogue systems in the future.

\section*{Acknowledgments}
This work is supported in part by the Microsoft Research Asia Internship program and the Shenzhen University Society Zero Universe platform. 
Dr. Hao Liao and Dr. Jianxun Lian are the supervisors.

\bibliography{custom}

\clearpage   
\onecolumn   
\appendix

\section{Prompt Templates}
\label{sec:prompt_templates}

This appendix provides the full prompts used for the API track tasks and for the LLM-as-Judge evaluation.

\subsection{Task 1: Tool-Calling Prompt}
\begin{tcolorbox}[title=Task 1: Tool-Calling Prompt, colback=gray!5, colframe=black!40]
\textbf{Instruction} \\
You are an assistant in estimating function names and arguments given some dialogues 
in a video game world. You will need the following information to respond to the user's input. 

Steps:  
1. Read the dialogue and the target item.  
2. From the given function information, select the functions that can obtain the information you need.  
3. Fill in the arguments needed by the function as appropriate.  

Note: You may select multiple functions or no functions at all.  

\textbf{Additional Information:} \{...\}  

\textbf{Dialogue:} The user input for the current turn is as follows.
\end{tcolorbox}

\subsection{Task 2: Dialogue Generation Prompt}
\begin{tcolorbox}[title=Task 2: Dialogue Generation Prompt, colback=gray!5, colframe=black!40]
\textbf{Instruction} \\
You are an assistant that plays the role of a character in a video game. Use the following character settings and knowledge to create your response.  

\textbf{Character Settings:} \{...\}  

\textbf{Knowledge:}  
- Knowledge from Function Calls: \{...\}  
- General Knowledge of All Items: \{...\}  

\textbf{Worldview:} \{...\}  
\end{tcolorbox}

\subsection{LLM-as-a-judge Reward Prompt}
The LLM-as-Judge evaluates the quality of role-playing dialogue based on four key dimensions:  
Scenario Adherence \& Quest Progression, NPC Believability \& Engagement, Persona Consistency, and Dialogue Flow \& Coherence.

\begin{tcolorbox}[title=LLM-as-a-judge Reward Prompt, colback=gray!5, colframe=black!40]
You are a professional dialogue critic in role-playing games. Your task is to evaluate a model-generated NPC response given a role-playing instruction and context.  

You must assess the response based on the following four dimensions:

1. \textbf{Scenario Adherence \& Quest Progression}  
- Does the NPC respond appropriately to the situation and task at hand?  
- Does it help move the current quest, story, or dialogue forward?  

2. \textbf{NPC Believability \& Engagement}  
- Does the response feel natural, immersive, and emotionally appropriate?  
- Is the response engaging and does it maintain conversational flow?  

3. \textbf{Persona Consistency (NPC Only)}  
- Is the NPC’s behavior, knowledge, and style consistent with their defined background and personality?  
- Are they saying things that fit their role, identity, and motivations?  

4. \textbf{Dialogue Flow \& Coherence}  
- Is the response well-structured, logically coherent, and contextually relevant?  

\textbf{Evaluation Instructions:} Output the result in XML format:  
\verb|<reason>...</reason>|  
\verb|<score>[0-5]</score>|  
\end{tcolorbox}

\begin{table}[H]   
\centering
\begin{tabular}{cl}
\hline
Score & Description \\
\hline
0 & Completely fails; irrelevant, incoherent, unfit for scenario. \\
1 & Very weak; major breaks in persona or flow. \\
2 & Partial attempt; notable flaws in tone or engagement. \\
3 & Reasonably coherent; minor lapses, lacks depth. \\
4 & Strong; consistent, engaging, with minor issues. \\
5 & Excellent; immersive, rich, convincingly in character. \\
\hline
\end{tabular}
\caption{Scoring criteria for LLM-as-a-judge evaluation.}
\end{table}

\end{document}